# EGFI: Drug-Drug Interaction Extraction and Generation with Fusion of Enriched Entity and Sentence Information

Lei Huang[1], Jiecong Lin[1], Xiangtao Li[3], Linqi Song[1], Zetian Zheng[1] and Ka-Chun Wong[1, 2, *]

[1] Department of Computer Science, City University of Hong Kong, Hong Kong SAR,

[2] Hong Kong Institute for Data Science, City University of Hong Kong, Hong Kong SAR and

[3] School of Artificial Intelligence, Jilin University, China.

*To whom correspondence should be addressed.

## Abstract

**Motivation:** The rapid growth in literature accumulates diverse and yet comprehensive biomedical knowledge hidden to be mined such as drug interactions. However, it is difficult to extract the heterogeneous knowledge to retrieve or even discover the latest and novel knowledge in an efficient manner. To address such a problem, we propose EGFI for extracting and consolidating drug interactions from large-scale medical literature text data. Specifically, EGFI consists of two parts: classification and generation. In the classification part, EGFI encompasses the language model BioBERT which has been comprehensively pre-trained on biomedical corpus. In particular, we propose the multihead self-attention mechanism and pack BiGRU to fuse multiple semantic information for rigorous context modeling. In the generation part, EGFI utilizes another pre-trained language model BioGPT-2 where the generation sentences are selected based on filtering rules.
**Results:** We evaluated the classification part on "DDIs 2013" dataset and "DDTs" dataset, achieving the F1 scores of 0.842 and 0.720 respectively. Moreover, we applied the classification part to distinguish high-quality generated sentences and verified with the existing growth truth to confirm the filtered sentences. The generated sentences that are not recorded in DrugBank and DDIs 2013 dataset also demonstrated the potential of EGFI to identify novel drug relationships.
**Contact:** lhuang93-c@my.cityu.edu.hk.
**Supplementary information:** Supplementary data are available at *Bioinformatics* online.

## 1 Introduction

There is enormous biomedical knowledge buried in the rapidly growing literature. However, those literature data are unstructured and heterogeneous. The cost of the manual annotation is expensive since it requires biomedical human experts. Fortunately, the recent advances in machine learning technologies have alleviated the problem; for instance, we can apply machine learning to extract meaningful biomedical information automatically (Wang et al., 2018; Pliner et al., 2019). Such meaningful information could be integrated for downstream clinical studies and biomedical engineering (Alobaidi et al., 2018). Moreover, machine learning models can even learn the potential relations between biomedical texts and discover novel knowledge among them (Sang et al., 2018; Lee et al., 2020), which could be confirmed with the downstream wet lab experiments.

Drug-drug interactions (DDIs) are the central annotations in precision medicine, given its overwhelming knowledge and available information (Herrero-Zazo et al., 2013; Zhang et al., 2020). The interactions of drugs may induce adverse reactions such as drug therapeutic failure (Franceschi et al., 2004); it could increase the treatment efficacy such as a complete resolution of pain of trigeminal neuralgia (Siniscalchi et al., 2011). Although the knowledge databases such as DrugBank (Wishart et al., 2018), ChEMBL (Gaulton et al., 2017) and STITCH (Kuhn et al., 2007) have already integrated the drug information such as DDIs, there are still potentially hidden and novel relationships to be discovered from the available literature. It is unrealistic to manually sort out and mine in the literature case by case. Therefore, the information extraction and generation technology could accelerate this process.

Recently, deep neural network models have become mainstream methods for extracting DDIs (Hong et al., 2020; Zhao et al., 2016; Sahu and Anand, 2018). In particular, the RNN and CNN models were often involved to learn the semantic and positional information of sentences and the related entities. Those deep neural network models could capture semantic information and outperform the traditional machine learning models such as SVM. However, a large amount of data is needed to train those deep neural network models to achieve robust performance while it is still very costly and time-consuming to annotate the data. Although RvNN (Lim et al., 2018) and latent tree learning12 have been proposed to learn syntactic information, the added features are still unable to address the original data scarcity.

Alternatively, transfer learning is a promising direction to enrich data features from external knowledge (Tan et al., 2018); for instance, BERT (Devlin et al., 2018) is a pretrained language model which has been already applied to many NLP tasks including relation extraction and achieved good performance. In essence, BERT makes use of the bidirectional transformer which can capture richer context than word2vec (Mikolov et al., 2013) and glove (Pennington et al., 2014). In particular, BioBERT is one of the BERT pretrained models for biomedical text mining (Lee et al., 2020). However, the solely fine-tuning on BioBERT may not be suitable for a particular dataset because the data distributions of the pretrained corpus and target corpus are different. In addition to the pretrained language models, data augmentation technology such as synonym replacement and random insertion can also be incorporated to improve the robustness (Wei and Zou, 2019). Text generation could also be regarded as one of the text augmentation technologies (Papanikolaou and Pierleoni, 2020). GPT-2 is also a pretrained language model; however, it is different from BERT. GPT-2 only encompasses the encoder of transformer and generates one token at a time. Therefore, it is intentionally designed for text generation. In addition, scholars also found that fusion of entity information including boundary and type is important for relation extraction (Zhong and Chen, 2020).

To address these issues, we propose a machine intelligence framework called EGFI, which is based on BioBERT that fuses multiple sentence and entity information to distinguish the interactions of the pairs of drugs in the sentences and incorporates BioGPT-2 which is inspired by BioBERT to generate text in a robust manner. Firstly, EGFI enriches the entity information by adding its boundary and type information. In classification part, the enriched sentences are fed into BioBERT to achieve the sentence sequence representation and semantic presentation. EGFI is then designed to rely on multihead self-attention mechanism and BiGRU to adapt the model to the target data. Finally, EGFI combines the semantic presentations of sentences, entity information produced by BiGRU and semantics presentation of sentences produced by BioBERT to learn context-rich information. In generation part, the enriched sentences are also fed into BioGPT-2. In order to generate authentic text, EGFI adopts the perplexity with the early stop mechanism to choose the best parameters. The generated sentences containing the boundary and type information of the drugs can be filtered to identify potential meaningful interactions by well-trained classification part. Our proposed EGFI framework facilitates with the strengths of BioBERT and BioGPT-2, in which BioBERT can learn the semantic representation of the input sentences and entities, and BioGPT-2 is suitable for text generation. Particularly, attention mechanism and pack BiGRU following BioBERT can learn the specific semantic information rather than fine-tuning BioBERT. The concatenation of the three semantic vectors enables the model to harness both entity information and sentence information in different learning stages. The well-trained EGFI can screen the generated data to select high-quality sentences which implies unrecorded promising interactions. The results show that EGFI can outperform others on the DDIs 2013 dataset and DTIs dataset for biomedical relation extraction. The text generated by EGFI can also find potential DDIs, which can be supported by other literature resources.

## 2 Dataset

### 2.1 DDIs 2013 dataset

In this study, we used DDIs 2013 dataset6. It is a manually annotated single sentence corpus consisting of 792 texts from DrugBank and 233 Medline abstracts. Among them, 18502 pharmacological substances and 5028 DDIs have been annotated. The pharmacological substances are classified into four types: drug (generic drug), brand (trade drug), group (drug classes), and drug-n (active substances not approved for human use). The DDIs are annotated from individual sentences with four relation types: mechanism (describing the way the interaction happens), effect (describing the consequence of interactions), advise (DDIs are described by a recommendation or advice), and int (DDIs are provided without any information). A fake type "negative" is also provided for neutral sentences.

The followings are the examples of the five types:

- Mechanism: Median gastric pH was significantly higher when **indinavir** was taken after **didanosine** administration.

- Effect: Synergism was observed when **GL** was combined with **cefazolin** against Bacillus subtilis and Klebsiella oxytoca.

- Advise: **Barbiturates** and glutethimide should not be administered to patients receiving **coumarin_drugs**.

- Int: **Chloral_hydrate** and methaqualone interact pharmacologically with orally administered **anticoagulant_agents**, but the effect is not clinically significant.

- NA: In vitro binding studies with human serum proteins indicate that **glipizide** binds differently than tolbutamide and does not interact with **salicylate** or dicumarol.

The statistics of DDIs 2013 are tabulated in Table 1. Numerically, 77% of the data are divided into a training dataset and the remaining data are considered as test data. In order to optimize the performance for each model, we randomly select 10% of training data to as the development dataset (or validation dataset) to tune the model parameters of interest in this study. Besides, the negative pairs are much more than positive pairs, leading to the data imbalance issue. Therefore, we employ text generation technology for data up-sampling on positive pairs and down-sampling on negative pairs, similar to the previous work (Hong et al., 2020; Sahu and Anand, 2018). Table 2 tabulates the data summary statistics after negative sample filtering.

**Table 1.** Statistics of DDIs 2013 dataset.

| Instances | DDIs type | Train | Test |
|---|---|---|---|
| Positive | Advise | 826 | 221 |
| | Effect | 1,687 | 360 |
| | Mechanism | 1,319 | 302 |
| | Int | 188 | 96 |
| Negative | | 23,665 | 4,712 |
| Total | | 27,685 | 5,691 |

**Table 2.** Statistics of DDIs 2013 dataset after negative sample filtering

| Instances | DDIs type | Train | Test |
|---|---|---|---|
| Positive | Advise | 814 | 221 |
| | Effect | 1,592 | 357 |
| | Mechanism | 1,260 | 301 |
| | Int | 188 | 92 |
| Negative | | 8,987 | 2,049 |
| Total | | 12,841 | 3,020 |

## 2.2 DTIs Dataset

The DTIs dataset was constructed by Hong et al (2020). It has more than 480k sentences from nearly 20 million PubMed abstracts. The labels of the sentences are chosen by aligning drug–target pairs against the DTI facts in DrugBank. The sentences are labelled with six types: substrate (the target acts upon the drug), inhibitor (drug that binds to the target impedes its function), agonist/antagonist (the drug that binds to the target activates or blocks its biological response), unknown (the interaction of drug–target pair is exited, but the action mechanism is not reported in DrugBank), other (all the other types of interactions with fewer occurrences), and the fake type "NA".

The followings are the examples of the five types:

- Substrate: Ketoconazole co-administration alters the pharmacokinetics of **praziquantel** in humans, possibly through inhibition of CYP3A, particularly **CYP3A4**, first-pass metabolism of praziquantel.

- Inhibitor: **Nintedanib** is a potent small-molecule, triple-receptor tyrosine kinase inhibitor (**VEGFR1**, VEGFR2, and VEGFR3; fibroblast growth factor receptor 1 [FGFR1], FGFR2, and FGFR3; and PDGFR\u03b1 and PDGFR\u00df).

- Agonist/Antagonist: A novel peptide, AMG 416 (formerly KAI-4169, and with a United States Adopted Name: **velcalcetide**), has been identified that acts as an agonist of the calcium-sensing receptor (**CaSR**).

- Unknown: We investigated the ability of **CobT** to act on either of two nitrogen atoms within a single, asymmetric **benzimidazole** substrate to form two isomeric riboside phosphate products.

- Other: Real-time RT-PCR showed that **ibuprofen** altered the expression of several genes including Akt, P53, PCNA, Bax, and **Bcl2** in the AGS cells.

- NA: Our results indicate that **PPA** on UPP is useful in assessing the grade of obstruction due to **BPE** when PFS fails, a finding still to be evaluated by prospective studies.

In order to compare the performance of EGFI with the existing methods, the data proportions of training dataset, validation dataset, and test dataset remain fixed. The statistics of DDIs 2013 are tabulated in Table 3.

**Table 3.** Statistics of DTIs dataset.

| Instances | DDIs type | Train | Validation | Test |
|---|---|---|---|---|
| Positive | Substrate | 1,710 | 12 | 18 |
| | Inhibitor | 2,612 | 20 | 19 |
| | Agnist/Antagonist | 855 | 11 | 10 |
| | Unknown | 2,534 | 37 | 26 |
| | Other | 604 | 3 | 10 |
| Negative | | 464K | 4,686 | 4,734 |
| Total | | 472K | 4,769 | 4,817 |

# 3 Method

## 3.1 The overview of EGFI

The structure of the automatic extraction and generation framework EGFI is outlined in Figure 1a. EGFI consists of two parts, classification and generation. Firstly, EGFI fuses the boundary and type of information to the sentence. In the classification part, the enriched sentences are fed into BioBert to capture the semantic information and sentence embedding, which is presented in Figure 1b. To transfer the model to the target data, we employ the multihead self-attention mechanism following pack BiGRU and fully connected layers to construct the deeply trained semantic representation of sentences and entities. We combine the CLS representation with the enriched semantic representation of input sentences and recognize the interaction types of the drug pairs by feeding them into a SoftMax classifier. In the generation part, the enriched sentences are fed into BioGPT-2 to generate the sentences. Then, the generated sentences are selected by the filtering rules. The filtered sentences are fed through the well-trained classifier (classification part) to become the high-quality sentences for literature supports. The confirmed high-quality sentences provide potentially meaningful interactions.

## 3.2 Input Representation

DDIs dataset is labelled at the single sentence level. Each sentence s has a pair of drugs (e1, e2) with an annotated relation {advise, effect, mechanism, int, negative}. Previous works adopted position embedding (Shi et al., 2018; He et al., 2018) or POS (part of speech) embedding (Hong et al., 2020; Hoesen and Purwarianti, 2018) to augment a neural network to learn the information of entities in each sentence. In this study, to capture the position and type information of drugs in each sentence based on BioGTP-2 and BioBERT, we insert a pair of special tokens '<e1i>' and '</e1i>' at the boundary of the first entity or drug where 1 denotes the first entity and i denotes the index of drug type {1: drug, 2: brand, 3: group, 4: drug_n}. At the boundary of the second entity, we insert a pair of special tokens '<e2i>' and '</e2i>' where 2 denotes the second entity; for example, the sentence with a pair of drugs<ibogaine, cocaine> after insertion special tokens will become:

"Only <e13> ibogaine </e13> enhances <e20> cocaine </e20> -induced increases in accumbal dopamine.

In this sentence, <e13> and </e13> indicate the location of "ibogaine" and we can infer from the second number in the special token that the drug type of ibogaine is a group. <e20> and </e20> also contains the location and type information of "cocaine". In addition to the insertion of special tokens, we also replace the entities with "drug1" and "drug2", avoiding the effect of different drug word embedding. The sentence presented above will become:

"Only <e13> drug1 </e13> enhances <e20> drug2 </e20> -induced increases in accumbal dopamine."

Given this preprocessed input sentence S = ($W_1$, ..., $W_n$) with the drug entities e1and e2, EGFI splits the sentence (except for the special tokens) into word pieces (a.k.a., subwords) by the WordPiece algorithm (Kudo and Richardson, 2018).

### 3.2.1 Text generation by BioGPT-2

GPT-2 (Radford et al.2019) is developed with deeper structures and more pre-trained textual data than the well-known GPT language model (Radford et al., 2018). They all utilize the decoder of the transformer and are pre-trained on a large amount corpus. Given k previous exit word tokens, the objectivity of both models is to predict the next word token and try to maximize the likelihood of the correct word token as follows:

$$L(U) = \sum_i \left( log P(u_i | u_{i-1}, \dots, u_{i-k}; \theta) \right) \quad (1)$$

where $\theta$ denotes the parameters of GPT-2.

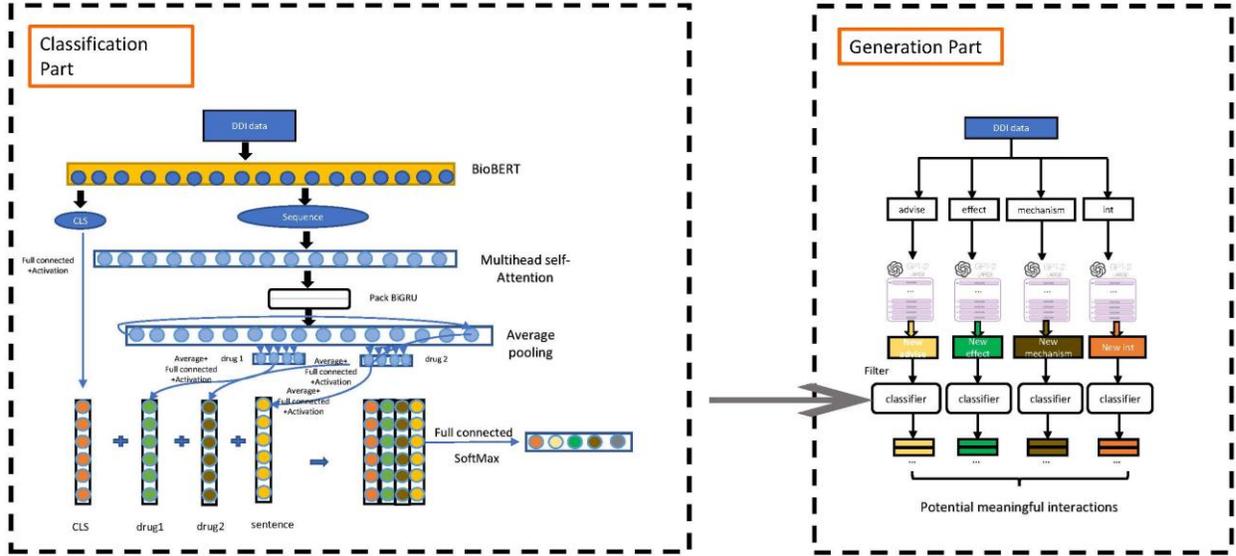

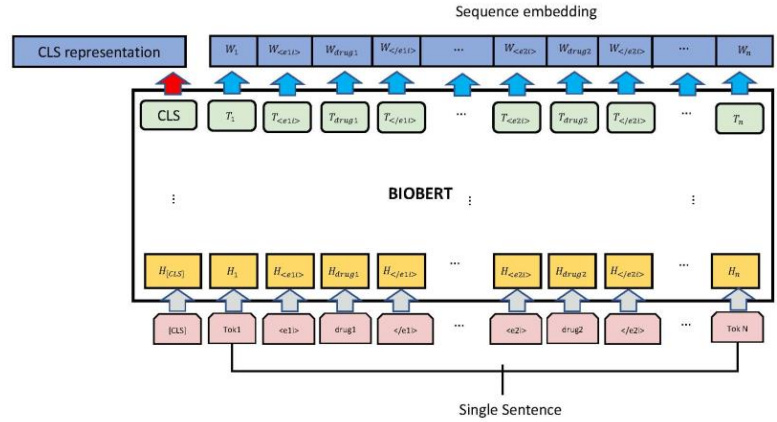

**Fig. 1.** The structure of EGFI. (1a) is the schematic structure of EGFI; (1b) illustrates the process of the single sentence feeding into BioBERT.

In this study, we used the GPT-2 pretrained on 500k PubMed abstracts with 744M parameters (Papanikolaou and ) which is called BioGPT-2. We fine-tuned four BioGPT-2 models on four types of sentences respectively. BioGPT-2 regards '<|endoftext|>' as the end token for each generated sentence. We only selected the generated data that each sentence only includes one pair of drugs with the special token that contains the type information. The type information should be correct according to the original data that have the drug type information.

**3.2.2 Semantic representation and sequence embedding by BioBert**

After the enriched sentences have been generated, EGFI inputs them into BioBERT. BioBERT is pretrained on PubMed abstracts for 1M steps, is similar to BioGPT-2. EGFI then converts the word pieces into a meaningful valued vector which is called the sequence output of BioBERT. Besides, the BioBERT part of EGFI also outputs the vector of CLS which represents the semantic meaning of the sentence.

**3.3 Capturing long and short dependency between words**

Human attention mechanism is derived from intuition. It is a strategy by which humans use limited attention resources to quickly filter out high-value information from high-volume data. The attention mechanism in deep learning draws on human attention thinking and is widely used in natural language processing (Hu, 2019). Words may have long or short dependencies with other words in one sentence. Therefore, EGFI adopts multi-head self-attention (Vaswani, 2017) to capture the hidden long and short dependency information among words.

For the multi-head attention layer, the sentence output of BioBert X consists of n word pieces which could be represented as follows:

$$X = (w_1, \ldots, w_n) \qquad (2)$$

where w denotes a vector with 768 word features. In order to learn the expression of multiple meanings, linearly mapping on X with different weights is computed to obtain three matrices:

$$
\begin{aligned}
Q &= Linear(X_{embedding}) \\
K &= Linear(X_{embedding}) \\
V &= Linear(X_{embedding})
\end{aligned} \qquad (3)
$$

Based on the three matrices, the attention score could be calculated by:

$$Attention(Q, K, V) = softmax\left(\frac{QK^T}{\sqrt{d_k}}\right) V \qquad (4)$$

where $\sqrt{d_k}$ turns the attention matrix into a standard normal distribution.

### 3.4 Semantic representation of sequence and entities after deep training

When training RNN (e.g., LSTM, GRU, and vanilla-RNN), it is difficult to batch the variable length of sequences. Therefore, it is necessary to pad the sentences to the same length for batch training. The padding sequences are meaningless. Pack BiGRU is designed to pack the padding input of sentences. EGFI only needs the words before padding tokens. Therefore, EGFI allows the words embedding before padding tokens to go through BiGRU. Figure 2 shows the processing example of pack BiGRU.

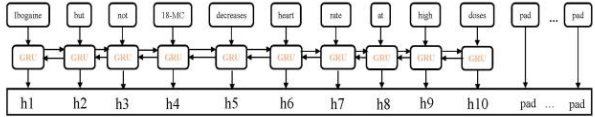

**Fig. 2.** Text processing example of pack BiGRU.

Besides, BiGRU could handle the ordered sequence from both backward and forward states. Given a sequence $X' = (e_1', \ldots, e_n')$ after BioBERT and multi-head attention, the output of BiGRU is denoted by:

$$H = (h_1, h_2, \ldots, h_n) \qquad (5)$$

Specifically, EGFI concatenates the first hidden state and the final hidden state by adding a fully connected neural network to construct the semantic representation of each sentence:

$$R_s = W_s \left[\tanh(concat(h_1, h_n))\right] + b_s \qquad (6)$$

For the semantic representation of entities, we suppose that $h_g$ to $h_i$ denote the hidden states of pack BiGRU for entity1 while $h_k$ to $h_m$ denote the hidden states of pack BiGRU for entity2. EGFI utilizes the same representation process as the previous work (Wu and He, 2019) which applies the average operation to construct the vector representation for each of the two target entities, following the fully connected work with an activating function (tanh).

$$E_1 = W_{e1}\left\{\tanh\left(\frac{1}{i-g+1}\sum_{t=g}^{i} h_t\right)\right\} + b_{e1} \qquad (7)$$

$$E_2 = W_{e2}\left\{\tanh\left(\frac{1}{m-k+1}\sum_{t=k}^{m} h_t\right)\right\} + b_{e2} \qquad (8)$$

### 3.5 Classification and optimization

EGFI adds the same fully connected layer for CLS representation of BioBERT $R_{cls}$ to result in the vector:

$$R_{cls} = W_{cls}\left[\tanh(r_{cls})\right] + b_{cls} \qquad (9)$$

Then EGFI concatenates the sequence representation vector of BioBERT $R_{cls}$, entity representation vectors $E_1$, $E_2$, and sequence representation vector of pack BiGRU $R_s$. After that, we calculate the possibility of the relation between a pair of entities using a SoftMax classifier:

$$M = [R_{cls}, E_1, E_2, R_s] \qquad (10)$$

$$P(r|M) = softmax(W_m M + b_m) \qquad (11)$$

EGFI uses the cross-entropy loss as the objective function to minimize:

$$J = -\frac{1}{N}\sum_{i=1}^{N} t_i \log(P(r|M)) \qquad (12)$$

where $t_i \in \{0, 1\}$ denotes the indicator variable if the relation is the same as the relation of samples and N is the number of relations. We minimize the loss by AdamW optimizer (Loshchilov and Hutter, 2018).

In order to solve the imbalanced data issue, we added weights for each relationship:

$$J_{weights} = -\frac{1}{N}\sum W_i t_i \log(P(r|M))_{i=1}^{N} \qquad (13)$$

$$W_i = \frac{\frac{C_{all}}{C_i}}{\sum_i \frac{C_{all}}{C_i}} \qquad (14)$$

where $C_{all}$ is the total number of the sentences and $C_i$ is the number of sentences with $i^{th}$ relation.

## 4 Experiments

For text generation stage, we propose a linear warm-up strategy to adjust the learning rate dynamically. Specifically, the learning rate is increased linearly and then decreased linearly (Figure 3). Such a warm-up strategy could avoid the learning curve oscillations given the large learning rate at the beginning.

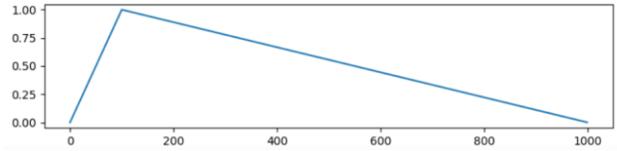

**Fig. 3.** The changing process of learning rate.

We split the original dataset into four data subsets; each of which only contains the sentences with one of the relations. Then, we fine-tune BioGPT-2 on each subset for five epochs and generate new sentences by the fine-tuned models. In order to result in the optimal model, we adopted the early stop mechanism, which we set the threshold of 5 continuously rising loss. We adopted the following strategies to filter out the sentences which are not applicable for model building:

- The sentences are too small. (less than 5 words).
- The sentences do not have any special token.
- The sentences only annotate one drug.
- The sentences annotate more than two drugs.

**Table 4.** The parameters setting of EGFI.

| Parts of EGFI | Parameters | DDIs type |
|---|---|---|
| Generation part | Learning rate | 3e-5 |
| | Warmup steps | 300 |
| | Batch size | 16 |
| | Maximum epoch | 5 |
| Classification part (DDIs 2013 dataset) | Leaning rate | 3e-5 |
| | Warmup steps | 0 |
| | Batch size | 8 |
| | Maximum epoch | 7 |
| | H in multi-head attention | 8 |
| | Dropout Probability of pack BiGRU | 0.5 |
| | Dropout Probability of FC layers | 0.1 |
| Classification part (DTIs dataset) | Leaning rate | 1e-5 |
| | Warmup steps | 0 |
| | Batch size | 8 |
| | Maximum epoch | 5 |
| | H in multi-head attention | 8 |
| | Dropout Probability of pack BiGRU | 0.5 |
| | Dropout Probability of FC layers | 0.1 |

**Table 5.** The results on DDI 2013 dataset

| Methods | P (%) | R (%) | $F_1$ score (%) |
|---|---|---|---|
| SCNN (Zhao et al., 2016) | 69.1 | 65.1 | 67.0 |
| CNN-bioWE (Liu et al., 2016) | 75.7 | 64.7 | 69.8 |
| Joint AB-LSTM (Sahu and Anand, 2018) | 73.4 | 69.7 | 71.5 |
| Position-aware LSTM (Zhou et al., 2018) | 75.8 | 70.4 | 73.0 |
| BERE (Hong et al., 2020) | 76.8 | 71.3 | 73.9 |
| CNN-focal loss (Sun, 2019) | 77.3 | 73.7 | 75.4 |
| Atten-BiLSTM (Zheng et al., 2017) | 78.4 | 76.2 | 77.3 |
| BioBERT (Peng et al., 2019) | ------ | ------ | 79.9 |
| Multiple-entity-aware-atten-BioBERT (Zhu et al., 2020) | 81.0 | 80.9 | 80.9 |
| EGFI | **83.4** | **85.0** | **84.2** |

The special tokens of each drug in the sentence are not correct.

This is an example of generated sentences:

  "<e10> Aspirin </e10> and insulin may enhance the antidiabetic action of <e22> antidiabetic_drugs </e22>."

For the classifier stage, by contrast, we padded every sentence to a large length, (300 tokens) which are larger than all the sentences for batch calculations. We also adopted the warm-up strategy similar to BioGTP-2. In addition, we added dropout layers for the pack BiGRU and the fully connected layers of $R_s$, $E_1$, $E_2$, and $R_{cls}$ for model regularization, avoiding overfitting. In the DDIs 2013 dataset, there are only training dataset and test dataset. In order to choose the optimal parameters, we split the validation dataset from the training dataset, which has the same size as the test dataset. The learning rate in the training process was fine-tuned on the validation (development) set using a grid search among {1e-5, 2e-5, 3e-5, 4e-5, 5e-5}. Table 4 outlines the optimal parameters setting in details. After that, we train the model on the original training dataset.

## 5 Result

We benchmarked EGFI on the test dataset of DDIs 2013 dataset to compare the performance of EGFI with other state-of-the-art models. we evaluated EGFI by micro-averaged F_1 score which is also adopted by the state-of-the-art models.

$$P = \frac{\sum_{n=1}^{N} TP_n}{\sum_{n=1}^{N} TP_n + \sum_{n=1}^{N} FP_n} \qquad (15)$$

$$R = \frac{\sum_{n=1}^{N} TP_n}{\sum_{n=1}^{N} TP_n + \sum_{n=1}^{N} FN_n} \qquad (16)$$

$$F_1 = \frac{2PR}{P+R} \qquad (17)$$

where $TP_n$, $FP_n$ and $FN_n$ denote the true-positive, false-positive, and false-negative instance numbers of the $n^{th}$ class respectively.

Table 5 shows the performances of EGFI and other models for DDI extraction on DDI 2013 dataset. Precision (P), Recall (R), and F_1 score of EGFI reach 84.2%, 83.5%, and 83.9% respectively, outperforming other baseline methods. Comparing with the models based on CNN and RNN, EGFI makes use of the pre-trained model to acquire rich representation of sentences and entities. In particular, it achieved noticeable performance improvement (77.3% vs 83.9%). More than that, EGFI fuses the position and type information of entities and enriched representation of sentences to enable its classifier part to learn as much knowledge as it can. It is quite different from the previous methods based on BERT.

EGFI has different performance in each type as shown in Table 6. EGFI performs the best in drugs with "advise" interaction while it has the lowest performance for the "int" relation. We reason that it is because of the scarce amount of "int" data to learn sufficient information to classify. It is also interesting that, though the amount of "advise" data is not the biggest among the four types, the performance of EGFI in "advise" is still the best.

**Table 6.** The performance of EGFI in each type.

| Type | P (%) | R (%) | $F_1$ score (%) |
|---|---|---|---|
| Advise | 86.7 | 88.7 | 87.7 |
| Effect | 78.1 | 92.7 | 84.8 |
| Mechanism | 89.5 | 85.5 | 87.2 |
| Int | 79.3 | 45.7 | 57.9 |

In order to understand the effect of each entity strategy on the performance of EGFI, we removed parts of EGFI to observe the corresponding decrease in $F_1$ score as shown in Table 7. We could find that, if we remove any part of EGFI, the performance is always dropped. It implies the methodology non-redundancy has been ensured in EGFI.

**Table 7.** The performance of EGFI without different parts.

| Enrich Strategy | P (%) | R (%) | $F_1$ score (%) | $\triangle$ |
|---|---|---|---|---|
| EGFI | 83.4 | 85.0 | 84.2 | —— |
| Non entity representation | 83.1 | 81.3 | 82.1 | 2.14% |
| Non sentence representation | 81.0 | 80.2 | 80.6 | 3.93% |
| Non self-attention | 74.4 | 86.5 | 80.0 | 4.64% |
| Non pack BiGRU | 81.4 | 79.9 | 80.7 | 3.81% |

In addition to DDIs 2013 dataset, researchers also applied the distant supervision method to develop a drug to target interaction dataset (DTIs) in which the interactions of drug and target are supported by bags of sentences. To better verify the extension ability of EGFI, we compared the classification part of EGFI with other baseline models on DTIs dataset. We used the same parameters as DDIs 2013 dataset. Table 8 shows the performances of EGFI and other baseline models. Comparing with other exited methods, EGFI achieves the highest F1 score and AUPRC.

**Table 8.** The results on DTIs dataset.

| Methods | $F_1$ score (%) | AUPRC (%) |
|---|---|---|
| BERE-AVE (Hong et al., 2020) | 46.0 | 38.4 |
| BERE-POOL (Hong et al., 2020) | 57.9 | 51.7 |
| BERE (Hong et al., 2020) | 62.5 | 52.4 |
| EGFI | **71.2** | **58.1** |

## 6 Applications to Identify Potential DDIs

As we mentioned before, EGFI can generate the sentences with the special tokens and relations. It can directly acquire the drugs and their interactions without any named entity recognition technology. Indeed, there are sentences generated by EGFI which could be meaningful since they do not exist in DDIs 2013 dataset. In this section, we used a case study to illustrate that EGFI can help researchers to identify potential DDIs that were not reported in DDIs 2013 dataset and DrugBank.

We first applied the generation part (BioGPT-2) of EGFI to generate the sentences for each type and selected the sentences which are not added to the original data for training. Then, we adopted the interaction extraction part of the well-trained EGFI to identify the type of generated sentences and check whether the generated sentences make biological sense or not. To select the most representative sentences, we ranked the sentences of each relation except "int" according to the softmax scores calculated by EGFI. Table 9 tabulates the five sentences with the highest scores in the relation "advise". The generation part of EGFI generated the sentences and annotated the drugs automatically.

We find that, the advice for drug_1 and drug_2 could be directly supported by the sentences. The keywords and cue phrases such as "consider", "avoid", "caution should be exercised", and "should be approached with caution" shows that EGFI could identify the important parts of the sentence to help the classification part to extract the relationships. In addition to examining the ability of EGFI in interaction type extraction, we further validate each advice more rigorously by inspecting the sentences and online resources. Among those advices generated and identified by EGFI, we found that the interaction of zalcitabine and amphotericin_b could not be found in DrugBank or DDIs 2013 dataset. However, other online resources such as antimicrobe database provide the related evidence. Specifically, the first sentence in Table 9 illustrates that we need to be cautious if we take zalcitabine and amphotericin_b at the same time or take the latter within a few hours after taking the former. DrugBank does not record the interaction between zalcitabine (accession number: DB00943) and amphotericin_b (accession number: DB00681). However, the sentence in the antimicrobe website[1] supports this advice:

"Amphotericin B, foscarnet, and parenteral aminoglycosides may interfere with renal clearance of zalcitabine."

This sentence shows that zalcitabine has the efficacy of renal clearance. Amphotericin b may interfere with such efficacy. Therefore, patients should be careful if they take zalcitabine and amphotericin B together.

In addition to the sentences with "advise" relationship, the sentences for the effect relationship also suggest the consequence of the pair of drugs. Specifically, the second sentence in Table 10 shows the effect of benzodiazepines and alcohol. Benzodiazepines belong to a class of psychotropic drug which is used for sedation, hypnosis, and anxiolytic effect. Alcohol also has the sedative effect. However, DrugBank and original DDIs 2013 dataset do not offer information about the interaction of these two drugs. We referred to other literature materials to verify the interaction. We find that, the website of American addition center records the information about the interaction of benzodiazepines and alcohol[2]:

"According to data from SAMHSA, individuals abuse the two drugs together because this practice enhances the effects of at least one of the drugs."

---

**Table 9.** Five sentences with the highest scores in "advise" type generated and identified by EGFI.

| Sentences with the largest score (advise) | Drug_1 | Druge_2 |
|---|---|---|
| Caution should be exercised when </e10> **zalcitabine** </e10> is administered concomitantly with or within # hours of <e20> **amphotericin_B** <e20>. | zalcitabine (drug) | amphotericin_b (drug) |
| If </e10> **diflunisal** </e10> therapy is required during treatment with <e20> **cyclosporine _ pump_inhibitors** <e20>, a reduction in the dosage of the steroid should be considered | diflunisal (drug) | cyclosporine_pump_inhibitors (drug) |
| The use of <e11> **ellence** </e11> should be avoided in patients receiving <e20> **tricyclic_antidepressants** <e20> because of the potential for serious adverse reactions (including myocardial infarction, cerebral vascular accident, renal failure, and cardiac arrhythmia) associated with tricyclic_antidepressants | ellence (brand) | tricyclic_antidepressants (drug) |
| Caution should be used when </e10> **alosetron** </e10> is coadministered with any of the following drugs: cisapride, anesthetic _ drugs (such as isoflurane, terfenadine, thioridazine, and <e20> **diltiazem** </e20>), beta - blockers (such as metoprolol), or type_1c_antiarrhythmics (such as propafenone) | alosetron (drug) | diltiazem (drug) |
| The use of proton _ pump_inhibitors such as </e10> **ketoconazole** </e10> and erythromycin during <e20> **trastuzumab** <e20> treatment should be approached with caution, with patient education emphasizing the need for close monitoring of the liver function tests | ketoconazole (drug) | trastuzumab (drug) |

**Table 10.** Five sentences with the highest scores in "effect" type generated and identified by EGFI.

| Sentences with the largest score (effect) | Drug_1 | Druge_2 |
|---|---|---|
| As with most cns depressants, use of <e10> bupropion </e10> may produce additive cns depression when coadministered with alcohol, <e22> sedatives </e22> and other drugs known to produce cns depression. | bupropion (drug) | sedatives (group) |
| This study has shown that <e12> benzodiazepines </e12> may increase the sedative effects of <e22> alcohol </e22>. | benzodiazepines (benzodiazepines) | alcohol (group) |
| Clinical studies, as well as postmarketing observations have shown that <e12> nsaids </e12> can reduce the natriuretic effect of <e22> thiazide_diuretics </e22>. | nsaids (group) | thiazide_diuretics (group) |
| <e12> anticholinergics </e12> may enhance the cns depressive effects of alcohol, <e22> sedatives </e22>, hypnotics, or other psychotropic _ medications | anticholinergics (group) | sedatives (group) |
| <e10> lofexidine </e10> and other benzodiazepines may decrease the effectiveness of oral contraceptives, certain antibiotics, <e22> quinidine </e22>, anticoagulants, and thyroid products. | lofexidine (drug) | quinidine (group) |

This sentence illustrates that the effect of the two drugs will be enhanced if we use them together. In addition, researchers also found that benzodiazepines and alcohol could be combined to achieve the addictive interactions (Linnoila, 1990). It could also verify this sentence to a certain extent.

Similarly, the sentences generated with mechanism relation illustrate how the drugs interact. The first sentence is Table 11 describes the way that erythromycin and antacids containing aluminum hydroxide interact. Erythromycin is a class of antibiotic which is used to treat bacterial infections[3]. Aluminum hydroxide is a class of antacids and is available in OTC medicines, having the efficacy of acid indigestion. We find that DrugBank and DDIs 2013 do not record the interaction of the two drugs. We also referred to other literature materials to verify the interaction. Some scholars found that the antacids studied including aluminum hydroxide would lead to the retardation of the dissolution of erythromycin (Arayne and Sultana, 1993). More than that, it is not advised to take the antacids after taking the erythromycin tablet within two hours[4], which could prove that the interaction of the two drugs would not be beneficial for their treatment. From Contemporary Clinic website, we find the evidence to support the validity of the generated sentence. It records that antacids containing aluminum, calcium and magnesium may decrease absorption of antibiotics, leading to the decrease of its efficacy[5].

The three examples of generated sentences verified above not only imply that, EGFI could extract the relationships from unstructured literature automatically and accurately; EGFI could help the researchers to identify the knowledge hidden in the literature. The examples also suggest the mining ability of EGFI, offering the potential DDIs. Besides, we also found that some interactions of drugs are not recorded in DrugBank and original

---

**Table 11.** Five sentences with the highest scores in "mechanism" type generated and identified by EGFI.

| Sentences with the highest score (mechanism) | Drug_1 | Druge_2 |
|---|---|---|
| The oral absorption of <e10> erythromycin </e10> was significantly reduced when given with antacids containing <e20> aluminum hydroxide </e20>. | erythromycin (drug) | aluminum hydroxide (drug) |
| Protease _ inhibitors: amprenavir, lopinavir, nelfinavir, and <e10> ritonavir </e10> have been shown to decrease plasma levels of <e20> saquinavir </e20>. | ritonavir (drug) | saquinavir (drug) |
| Coadministration with oral anticoagulants (warfarin, <e10> phenytoin </e10>, propranolol, etc.) significantly decreased the steadystate plasma <e20> tolbutamide </e20> concentrations. | phenytoin (drug) | tolbutamide (drug) |
| Other known cyp3a4 inhibitors such as <e10> quinidine </e10>, fluoxetine, and paroxetine have been shown to increase the plasma concentration of <e20> saquinavir </e20> significantly when coadministered with these agents. | quinidine (drug) | saquinavir (drug) |
| However, other agents that are potent inducers of cyp3a4 (e.g., phenobarbital, phenytoin, dexamethasone, and <e10> carbamazepine </e10>) may decrease plasma <e20> saquinavir </e20> concentrations. | carbamazepine (drug) | saquinavir (drug) |

dataset. Even worse, they are not supported by other online resources and literature neither. In the future, we could conduct web lab experiments to confirm the potentially novel DDIs, which could provide meaningful insights in the future.

# 7 Error Analysis

Similar to the work of Zhu et al. (2020), we presented a confusion matrix for error analysis (Figure 4a). The color intensity denotes the proportion of the relation. We also followed the normalization process of the previous work (Figure 4b), which help us observe the misclassified proportion clearly. From these two figures, we could summarize the types of errors:

- All positive relations have samples to be misclassified into negative relations.

- Many samples of "int" relation are misclassified into "effect" relation.

- Some samples of "negative" are misclassified into "effect" relation.

- The proportions of "mechanism" samples misclassified to "negative" relations and "int" samples misclassified to "negative" relations are relatively large.

Our results are similar to the previous work (Sahu and Anand, 2018; Zhu et al., 2020). First error indicates that the classifier cannot distinguish negative and positive samples as it confuses some samples of all positive relations with negative relations. The number of negative training samples (8987) is much larger than positive training samples (3854). Such a data imbalance issue could be considered as the reason for the first error. For the second error, we think the small number of "int" relations should be the reason for its misclassifications There are only 188 samples of "int" relations which are much less than other relations. Although we adopt the weights that are based on the number of samples to alleviate the unbalanced dataset effect, the result suggests that we should take measures to improve the performance. Besides, the semantics and structures of some samples of "int" are similar to those of "effect" samples. However, the interesting point is that 37 samples of "int" are misclassified into "effect" relations while no effect sample is misclassified into "int" relations as shown in Figure 4a. It reflects that EGFI tends to treat "int" samples as

"effect" samples. For example, EGFI misclassified a sentence of "int" relation ("other drugs which may enhance the neuromuscular blocking action of <e10> drug1 </e10> such as mivacron include certain <e22> drug2 </e22>") into "effect" relation since the word "enhance" in this sentence may confuse EGFI that the drug1 and drug2 has effect relation. It suggests the semantic similarity between the sentences of "effect" and the sentences of "int" relations. For error3, some negative samples are also misclassified into "effect" relations, which means that the semantics and structures of effect relation are complicated and confusable. For the last error, EGFI confuses some "mechanism" samples and "int" samples with "negative" samples, also reflecting its semantic similarity.

**4a**

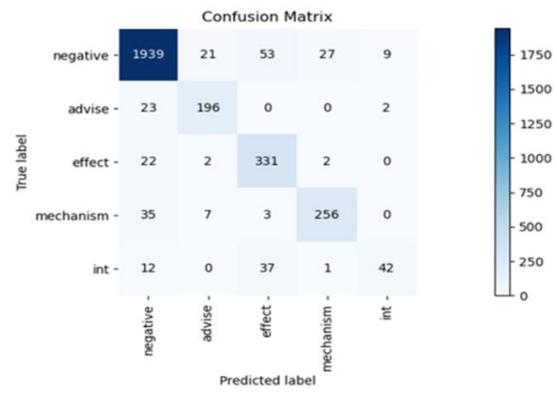

**4b**

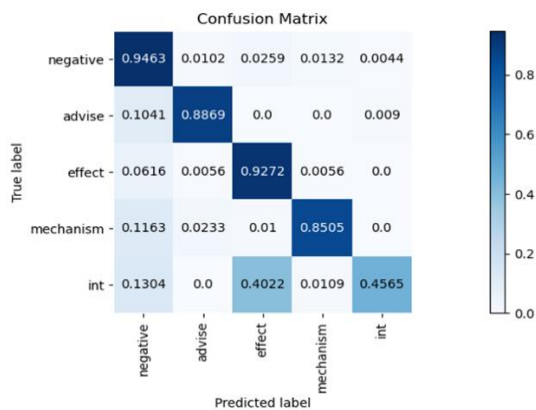

**Fig. 4.** Confusion Matrix. (3a) is the confusion matrix without normalization; (3b) is the normalized confusion matrix.

## 8 Discussion

In this work, we proposed EGFI which is an intelligent composed machine learning framework for information extraction and generation for DDIs. The classification part of EGFI could automatically extract the interactions between drugs more accurately than other previous models. The results of the classification part show that our proposed incorporation of BioBERT can make use of a large amount of external knowledge, improving the performance on the DDIs 2013 dataset and DTIs dataset greatly. By using multi-head attention mechanism and packed BiGRU, EGFI can capture the short-distance and long-distance dependencies in different attention spaces and learned the entity and sentence information. The fusion of enriched sentences and entities information could help the model learn the local contextual features of the entities and global features of sentences. The promising accuracy of EGFI could provide reliable relation extraction ability for researchers to utilize a large number of raw texts. We also conducted the error analysis to elucidate the details of the misclassification since the unbalanced data can affect the performance of EGFI, given the textual ambiguity in the reality.

Interestingly, we even observe that the generation part of EGFI could generate sentences that contain the interactions which are not reported in DrugBank and DDIs 2013 dataset by making use of BioGPT-2. Those potential interactions could be verified with other online resources or wet-lab experiments. We speculate that wet-lab experiments could be conducted to find the new meaningful DDIs presented in the generated sentences. Such case studies demonstrate the promising function of EGFI, identifying the potential relationships of drugs.

Overall, EGFI could be used as an intelligent information extraction tool, helping the researchers to not only extract the exiting DDIs and but also mine for potential DDIs. As the future work, we could import the external information of drugs such as the molecular structure and contextual texts to improve the performance of EGFI.

## Acknowledgements

The work described in this paper was substantially supported by the grant from the Research Grants Council of the Hong Kong Special Administrative Region [CityU 11200218], one grant from the Health and Medical Research Fund, The Food and Health Bureau, The Government of the Hong Kong Special Administrative Region [07181426], and the funding from Hong Kong Institute for Data Science (HKIDS) at City University of Hong Kong. The work described in this paper was partially supported by two grants from City University of Hong Kong (CityU 11202219, CityU 11203520). This research is also supported by the National Natural Science Foundation of China under Grant No. 32000464.